\documentclass{article}

\usepackage[preprint]{style}

\usepackage[utf8]{inputenc} % allow utf-8 input
\usepackage[T1]{fontenc}    % use 8-bit T1 fonts
\usepackage[colorlinks=true]{hyperref}       % hyperlinks
\usepackage{url}            % simple URL typesetting
\usepackage{booktabs}       % professional-quality tables
\usepackage{amsfonts}       % blackboard math symbols
\usepackage{nicefrac}       % compact symbols for 1/2, etc.
\usepackage{microtype}      % microtypography
\usepackage{xcolor}         % colors

\usepackage{multirow}
\usepackage{graphicx}
\usepackage{placeins}

\usepackage{algorithm,algpseudocode}
\usepackage{wrapfig}
\usepackage{mathtools}
\usepackage{amssymb}
\usepackage{subcaption}
\usepackage{caption}

\newcommand{\conv}[1]{\operatorname{conv}{\left(#1\right)}}

\newcommand{\convth}[1]{\operatorname{conv}{_{k \times k}\left( #1 \right)}}

\newcommand{\concat}[1]{\operatorname{concat}{\left( #1 \right)}}

\title{Learning Compact Vision Tokens for Efficient Large Multimodal Models}

\author{%
  Hao Tang\thanks{Equal Contribution}\\
  Department of Computer Science\\
  Central South University\\
  \texttt{tanghao\_csu@csu.edu.cn} \\
  \And
  Chengchao Shen$^{*}$\thanks{Corresponding Author}\\
  Department of Computer Science\\
  Central South University\\
  \texttt{scc.cs@csu.edu.cn} \\
}

\begin{document}

\maketitle

\begin{abstract}
    Large multimodal models (LMMs) suffer significant computational challenges due to the high cost of Large Language Models (LLMs) and the quadratic complexity of processing long vision token sequences.
    In this paper, we explore the spatial redundancy among vision tokens and shorten the length of vision token sequences for inference acceleration. 
    Specifically, we propose a Spatial Token Fusion (STF) method to learn compact vision tokens for short vision token sequence, where spatial-adjacent tokens are fused into one.
    Meanwhile, weight-frozen vision encoder can not well adapt to the demand of extensive downstream vision-language tasks.
    To this end, we further introduce a Multi-Block Token Fusion (MBTF) module to supplement multi-granularity features for the reduced token sequence. 
    Overall, we combine STF and MBTF module to balance token reduction and information preservation, thereby improving inference efficiency without sacrificing multimodal reasoning capabilities.
    Experimental results demonstrate that our method based on LLaVA-1.5 achieves comparable or even superior performance to the baseline on 8 popular vision-language benchmarks with only 25\% vision tokens of baseline.
    The source code and trained weights are available at \url{https://github.com/visresearch/LLaVA-STF}.

\end{abstract}    
\section{Introduction}
\label{sec:intro}
Large multimodal models (LMMs)~\cite{llava,llava2} based on Large Language Models (LLMs)~\cite{gpt4,gemini,llama} have shown remarkable multimodal reasoning capabilities.
LMMs leverage a vision encoder such as CLIP-ViT~\cite{clip} to embed images into vision tokens as the prefix visual context,
and feed them into a large language model pretrained on large-scale text corpus.
Despite the impressive capabilities, LMMs face substantial computational challenges, which limit the scalability and efficiency.

% The first challenge, also the primary factor for the high computation cost, arises from the LLMs,
% since the vision encoder is usually relatively small. 
% The first factor for the high computation cost of LMMs is 
% The primary reason behind the high computation cost of LMMs is 
The high computation cost of LMMs primarily comes from LLMs, where vision encoder for LMMs is obviously smaller than the corresponding LLM.
For example, CLIP-ViT-L adopted LLaVA~\cite{llava,llava2} only has 0.3B parameters, 
while the corresponding LLM, such as LLaMA~\cite{llama} or Vicuna~\cite{vicuna}, contains 7B or 13B parameters.
Although using LLMs with fewer parameters, such as Phi-2~\cite{phi}, can alleviate this burden, it often leads to obvious performance drops on visual question-answering and reasoning. 

\begin{figure}[t]
	\centering
	\includegraphics[width=0.6\linewidth]{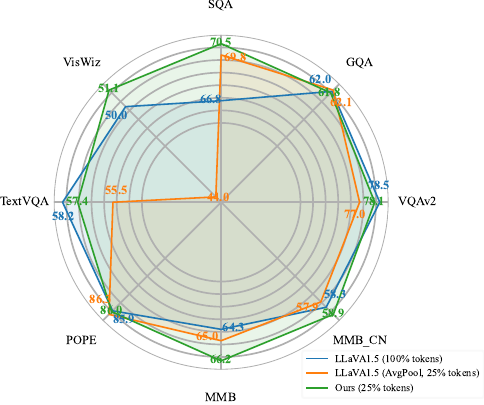}
	\caption{
		Spatial token redundancy in the visual context of LMMs.
		The performance gap between LLaVA-1.5-7B (AvgPool) with only 25\% vision tokens and the original LLaVA-1.5-7B is not so obvious, except VisWiz, demonstrating the excessive redundancy of vision tokens.
		% Our method combines STC and MLTC to reduce spatial vision redundancy, achieving comparable or even better performance than the original LLaVA-1.5-7B.
		Based on STC and MLTC, our method achieves comparable or even better performance than the original LLaVA-1.5-7B.
	}
	\vspace{-2em}
	\label{motivation}
\end{figure}

% Two factors affect the computation cost of LMMs, the number of input tokens and the parameters of LLM.
Another solution to improve inference efficiency of LMM is to reduce the number of vision tokens fed into LLM.
Since the number of vision tokens produced by vision encoder reaches hundreds even thousands, significantly surpassing the number of text tokens, the reduction of vision token can significantly improve the inference efficiency of LMM. 
As a result, several methods~\cite{tokenpacker,ge-mini,prumerge} are proposed to shorten the length of vision token sequence, which are fed into LLM.
However, the balance between the token reduction and information retention remains an open question.

In this paper, we explore the spatial redundancy present in the visual context of LMMs. 
As shown in Figure~\ref{motivation}, we retrain LLaVA-1.5-7B by simply reducing the number of vision tokens using average pooling before fed into LLM, where four adjacent tokens are averaged as one with a stride of 2, and only 25\% tokens are remained. 
The reduced model is designated as LLaVA-1.5-7B (AvgPool).
We surprisely find that its performance drop on several popular benchmarks is not obvious, except VisWiz~\cite{viswiz}.
Especially on POPE~\cite{pope} and GQA~\cite{gqa} benchmarks, LLaVA-1.5-7B (AvgPool) even outperforms the original LLaVa-1.5-7B.
The results support that excessive redundancy indeed exists in current LMMs.

% To tackle this problem, we propose a spatial token compression method,
% since image tokens within a small spatial window are highly similar in semantic information and thus often exhibit spatial redundancy.
To reduce the redundancy of vision tokens, we propose a \emph{Spatial Token Fusion} (\emph{STF}) method, which fuses adjacent vision tokens into one to shorten the token sequence.
Unlike previous methods, we address the spatial redundancy before vision tokens are fed into LLM. 
Instead of simply averaging adjacent vision tokens, our approach concatenates adjacent $k^2$ vision tokens in the sliding window with size of $k \times k$ along the channel dimension.
% Then, a linear projection is trained from scratch to fuse semantic information while bridging the vision and language models.
Then, we introduce learnable \emph{Spatial Token Fusion} (\emph{STF}) module to adaptively fuse features of adjacent $k^2$ vision tokens, while bridging the representations between vision encoder and LLM.
Compared to plain average pooling, our approach aims to preserve more information during redundancy reduction. 

Since vision encoder of LLaVA-style LMM is generally fixed during training, vision tokens generated by vision encoder can not well adapt to the demand of target tasks, especially for some tasks that require detail information of the give image.
To capture more detail visual information of image, we further propose a \emph{Multi-Block Token Fusion} (\emph{MBTF}) module to integrate low-level features with high-level semantic features, thus improving the compactness of the fused vision tokens. 
In this manner, our method can adaptively access multi-level features from vision encoder for widespread downstream vision-language tasks without the retraining of vision encoder.
Moreover, compared to LLM, the computation cost of additional modules, including STF and MBTF, can be ignored, yet the reduced sequence of vision tokens can significantly accelerate the inference of LMM.

In summary, the contributions of our method can be summarized as below.
\begin{itemize}
    \item We propose a Spatial Token Fusion module, which learns compact vision tokens to significantly shorten the vision token sequence fed into LLMs, thus effectively accelerating the inference of LMMs. 
    \item We propose a Multi-Block Token Fusion module to adapt the feature demand of extensive downstream vision-language tasks without the retraining of vision encoder. 
    \item Extensive experiments on LLaVA~\cite{llava} show that our approach achieves comparable or even superior performance to LLaVA-1.5-7B~\cite{llava2} on popular vision-language benchmarks with only 25\% vision tokens of the original one.
\end{itemize}

\section{Related Work}
\label{sec:related}
% todo

In this section, we briefly review related works about the acceleration of both large multimodal models (LMMs) and large language models (LLMs).

\subsection{Acceleration of LMMs}
Efforts to optimize the efficiency of LMMs have explored diverse strategies, 
such as lightweight vision encoders~\cite{vit,eva,llavanext,shen2023inter}, 
sparsely activated MoE architectures~\cite{moe}, 
and parameter-efficient language models~\cite{phi}. 
Among these, vision token pruning~\cite{ge-mini,llavauhd,texthawk,tang2025data} has gained traction due to its ability to shorten visual sequences without altering model parameters. 
For example, LLaVA-PruMerge~\cite{prumerge} merges redundant tokens at CLIP's penultimate layer, 
while FastV~\cite{fastv} employs adaptive attention patterns to prioritize essential tokens and prune others. 
Concurrently, LLaVolta~\cite{llavolta} introduces progressive token compression across training stages, 
balancing efficiency and performance.
TokenPacker~\cite{tokenpacker} proposes a coarse-to-fine visual projector that 
hierarchically compresses high-resolution image features through 
downsampling, point-region interaction, and cross-layer fusion 
to generate compact vision tokens.
TinyChart-3B's vision token merging~\cite{tinychart} dynamically reduces high-resolution input processing overhead 
by fusing similar vision tokens within each transformer layer.
YOPO~\cite{yopo} integrates three strategies, 
including 126 neighbor-aware vision token attention,
pruning of inactive visual attention heads, and selective layer dropping for visual computations, 
to improve the inference efficiency of LMMs. 
LLaVA-Mini~\cite{llavamini} introduces modality pre-fusion module to fuse vision tokens and text tokens for efficient inference.

In spite of encouraging performance achieved, the above methods require the involvement of text tokens to prune the number of vision tokens, thus compromising model performance due to the loss of fine-grained visual details.
In comparison, our method fuses adjacent vision tokens for information preservation, while adaptively accessing to features from different layers for widespread vision-language tasks.

\subsection{Acceleration of LLMs}
The quadratic computational complexity of transformer-based LLMs, 
which scales with the square of the input sequence length~\cite{attention}, 
has motivated substantial efforts to address inherent redundancy in these architectures. 
Prior research has explored two primary directions: parameter sparsification through weight pruning~\cite{mv,sparsegpt} and attention head reduction~\cite{sixteen}, 
and sequence compression to mitigate the overhead of long token sequences.
For the latter, hierarchical approaches like Pyramid Transformers~\cite{pyramid} progressively downsample token sequences across layers, 
while Nawrot et al.~\cite{et} propose adaptive sequence compression by semantic boundary prediction.
Recent VCC~\cite{vcc} further introduces layer-wise token aggregation by select important tokens.

However, the integration of visual encoders with LLM decoders introduces modality-specific computational bottlenecks, 
particularly in processing lengthy vision token sequences derived from high-resolution images. 
Unlike unimodal compression that prioritizes linguistic patterns, 
vision-language interactions demand modality-aware token reduction to preserve critical spatial-semantic correlations. 
To bridge this gap, we propose a novel vision token compression framework 
that strategically reduces the number of vision tokens fed into the LLM component of LMMs.

\begin{figure*}[t]
	\centering
	\includegraphics[width=\linewidth]{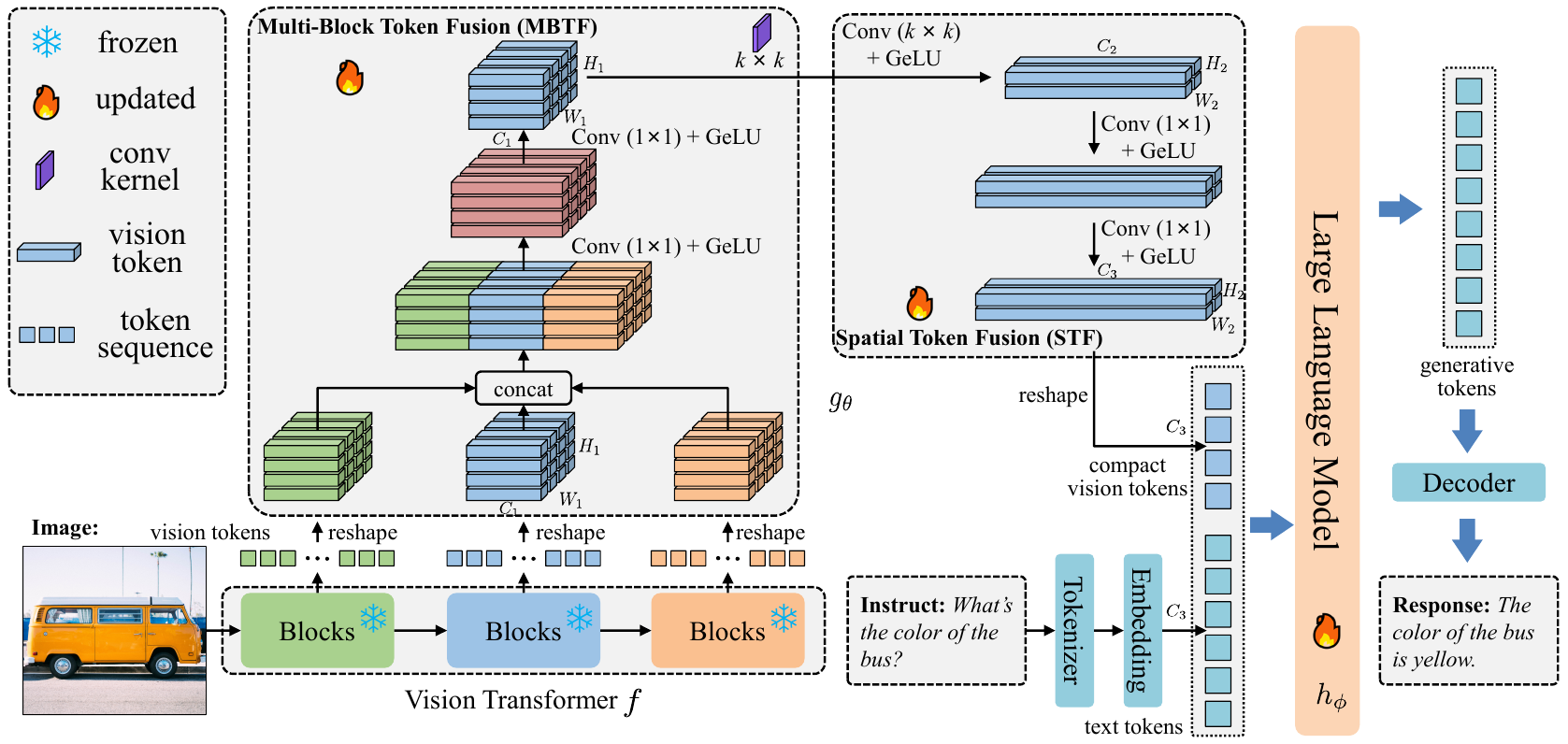}
    \caption{The overview of our method.
	Our method follows LLaVA-style architecture and introduces additional Multi-Block Token Fusion (MBTF) and Spatial Token Fusion (STF) modules.
	Based on basic LLaVA, we obtain vision tokens produced by selected intermediate blocks of vision encoder and then fuse them along the channel dimension for multi-granularity features. 
	Then, the fused vision tokens are fed into Spatial Token Fusion module, where convolution with kernel size of $k \times k$ is applied on the above vision tokens to aggregate vision tokens in $k \times k$ neighborhood and obtain more compact vision tokens, thus reducing spatial redundancy of input vision token sequence for large language model. 
	Finally, both compact vision tokens and text tokens from input instruct are fed into large language model to generate the corresponding response.
	}
	\vspace{-1em}
	\label{fig:overview}
\end{figure*}

\section{Method}

In this section, we first introduce the preliminaries of LLaVA-style large multimodal model. 
Then, we give a brief overview of our method. 
Afterward, we present our proposed Multi-Block Token Fusion module and Spatial Token Fusion in detail.
Finally, we depict the optimization of our method.

\subsection{Preliminaries}
Given an input image $X_v$, the vision tokens $X^l_v=f_l(X_v)$ is the output of the $l$-th block of the vision encoder $f$.
To bridge the gap between image and language modalities, LLaVA introduces a linear- or MLP-based projector $g_{\theta}$ ($\theta$ is the parameters of projector) to map the vision tokens $X^l_v$ into the text embedding space and obtains aligned tokens $Z_v = g_{\theta}(X^l_v)$, which have the same dimension as the text embedding in large language model $h_{\phi}$.
Then, aligned tokens $Z_v$ and instruct $X_{\rm instruct}$ are fed into the LLM $h_{\phi}$ to generate response as $X_{\rm response} = h_{\phi}(Z_v, X_{\rm instruct})$.

In this paper, we aim to reduce the number of aligned tokens $Z_v$ in a lossless manner to shorten the length of vision token sequence fed into LLM, thus accelerating the inference of LMM.
To this end, we focus on the design of projector $g_{\theta}$ in the following sections. 

\subsection{Overview}

The goal of token reduction is to minimize the length of vision token sequence, while maximizing the performance of the reduced large multimodal model.
To accomplish this goal on LMM, we explore the solution to reduce spatial redundancy among vision tokens, which stems from the local similarity property of natural images, especially for high-resolution images. 
Meanwhile, we also consider information preservation during redundancy reduction, thus minimizing the performance drop.

As shown in Figure~\ref{fig:overview}, we adopt a dual stage token fusion strategy to implement the projector $g_{\theta}$ between vision encoder and LLM.
In the first stage, we extract vision tokens from the selected blocks of vision encoder.
The extracted multi-block vision tokens are fused into compact one, which contains semantic representations with various granularities, thus preserving information for extensive downstream tasks. 
In the second stage, we further fuse the above compact tokens in the neighborhood to reduce the spatial redundancy.
After the above two token fusion steps, the fused vision tokens are further aligned to text embedding space of large language model.
Finally, combined with text tokens of instruct, the fused vision tokens are fed into LLM to generate the corresponding response in a significant efficient manner.

\subsection{Multi-Block Token Fusion}

In this section, we focus on the integration of multi-block tokens from vision encoder, to obtain multi-granularity representations.
Our main target is to learn compact vision tokens, which embrace multi-level semantic features of vision encoder, thereby adaptive to extensive vision-language tasks.

To fuse representative features from vision encoder $f$, we select vision tokens from $M$ blocks, whose indices can be denoted as $\{ l_i \}_{i=1}^{M}$.
Since the features from adjacent blocks have similar semantics, directly fusing of all these features would introduce substantial computation cost.
Hence, the indices $\{ l_i \}_{i=1}^{M}$ of selected blocks are evenly sampled. 
For LLaVA-1.5-7B, its vision encoder, ViT-L/14, has 24 blocks, and the indices of sampled blocks are $\{3, 6, 9, 12, 15, 18, 21, 24\}$. 

As depicted in Figure~\ref{fig:overview}, we introduce Multi-Block Token Fusion (MBTF) module to fuse multi-block tokens as follows.
First, we reshape the vision tokens, $\{ X^{l_i}_v \}_{i=1}^M$, as feature maps for better understanding of follow-up operation.
Then, the selected features are concatenated along channel dimension. 
Next, we fuse the concatenated features using two sequential convolution modules with kernel size of $1 \times 1$, followed by GeLU~\cite{hendrycks2016gaussian} activation function. 
To improve the compactness of the fused vision tokens, we progressively reduce the size of channel dimension in the convolution step.
For LLaVA-1.5-7B, the channel dimension size of the above convolution operations are 4096 and 1024, respectively, where the token dimension of vision encoder is 1024.
The overall process of MBTF can be presented as
\begin{equation}
\label{eq:mbtf}
	X_v^{\rm MBTF} = \conv{\conv{\concat{\{ X^{l_i}_v \}_{i=1}^M}}}, 
\end{equation}
where $\operatorname{conv}$ denotes the convolution with kernel size of $1 \times 1$, followed by GeLU.

\subsection{Spatial Token Fusion}
In this section, we further reduce spatial redundancy of the above multi-block fused vision tokens.

Let $H_1 \times W_1 \times C_1$ denotes the shape of multi-block fused vision tokens $X_v^{\rm MBTF}$, where $H_1$, $W_1$ and $C_1$ are the height, width and channel size of $X_v^{\rm MBTF}$, respectively. 
The text embedding of LLM is set to $C_3$.
Generally, the token dimension of LLM adopted in LMM is significantly larger than the one of the corresponding vision encoder, namely $C_3 \gg C_1$. 
An intuitive idea to shorten the length of vision token sequence is concatenating multiple vision tokens as one, whose dimension is close to the one of text token, thereby fusing multiple vision tokens without loss of information. 

As depicted in Figure~\ref{fig:overview}, we introduce Spatial Token Fusion module to reduce the redundancy of vision tokens. 
Specifically, we adopt convolution operation with kernel size of $k \times k$ to fuse the $k^2$ tokens of $X_v^{\rm MBTF}$ into a compact token as follows
\begin{equation}
    X_1^{\rm STF} = \convth{X_v^{\rm MBTF}},
\end{equation}
where $\operatorname{conv}_{k \times k}$ denotes the convolution operation with kernel size of $k \times k$ and stride size of $k \times k$, followed by GeLU.
The size of $X_1^{\rm STF}$ is $H_2 \times W_2 \times C_2$, where $H_2 = \frac{H_1}{k}$ and $W_2 = \frac{W_1}{k}$. 
If $\operatorname{conv}_{k \times k}$ is regarded as a learnable concatenation operation, we set $C_2 = k^2 \cdot C_1$.
For LLaVA-1.5-7B, the dimension of vision token and text token are 1024 and 4096, respectively. 
If we set $k=2$, the dimension of fused vision token $C_2$ is identical to the one of text token $C_3$ ($C_2 = C_3 = 4096$), thus achieving lossless token reduction.

For more general solution, adjacent $k^2$ vision tokens can be also fused to any number of tokens, by tensor reshaping.
Specifically, the fused token $X_1^{\rm STF} \in \mathbb{R}^{H_2 \times W_2 \times C_2}$ can be reshaped as $E$ tokens, where the shape of each token is $H_2 \times W_2 \times \frac{C_2}{E}$.
This provides more flexibility for spatial token fusion. 

To align the fused vision tokens with text embedding of LLM, we further introduce two additional convolution modules with kernel size of $1 \times 1$, followed by GeLU activation function and obtain aligned vision token sequence $X_v^{\rm STF} \in \mathbb{R}^{(\frac{H_1}{k} \cdot \frac{W_1}{k} \cdot E) \times C_3}$.
Finally, the number of aligned vision tokens fed into LLM, is reduced from $\left( H_1 \cdot W_1 \right)$ to $\left( \frac{H_1}{k} \cdot \frac{W_1}{k} \cdot E \right)$, where adjacent $k^2$ vision tokens are fused into $E$ tokens and $E < k^2$.

\subsection{Optimization}

Following the training scheme of LLaVA~\cite{llava,llava2}, we optimize our reduced LMM model by maximizing the probability $p$ of generating target response $X_r$ by
\begin{equation}
p(X_r | X_v, X_{\rm instruct}) = 
 \prod_{i=1}^N p_{\theta, \phi}(x_i | X_v^{\rm STF}, X_{\rm instruct}, X_{r, <i}),
\end{equation}
where $\theta$ and $\phi$ are learnable parameters of projector $g_\theta$ and LLM $h_\phi$, $N$ is the length of response $X_r$ and $X_{r, <i}$ are response tokens before the current prediction token $x_i$.

We also follow the two-stage optimization scheme of LLaVA, including feature alignment pretraining and end-to-end finetuning. 
In the pretrainig stage, only parameters $\theta$ are updated and others are fixed.
In the finetuning stage, only parameters $\phi$ are learnable and other parameters are frozen.

\section{Experiments}
First, we present our experiment setup, including model architecture, evaluation benchmarks and implementation details.
Afterward, we evaluate our models on extensive popular vision-language benchmarks and compare with other efficient LLaVA models.
Then, we conduct ablation studies to dissect the key components of our approach.
Finally, we give some cases of vision-language reasoning and compare our method and the baseline LLaVA.

\subsection{Experiment Setup}

\subsubsection{Model Architecture}
In this work, we mainly focus on the acceleration of LLaVA models~\cite{llava,llava2}.
Our models use the CLIP ViT-L/14~\cite{clip} as the vision encoder and the resolution of input image is set to $336 \times 336$.
Besides, we utilize Vicuna-1.5-7B~\cite{vicuna} as the LLM backbones.
The number of selected blocks $M$ for Multi-Block Token Fusion module is set to 8.
By default, the kernel size $k$ for Spatial Token Fusion is set to $2\times 2$ and the number of target fused tokens $E$ is set to 1.
To fully integrate information from different layers, we select every three layer from the 24 blocks of the vision encoder.
% The MLPs in the Spatial Visual Compressor and in the Cross-Layer Visual Connector both consist of two linear layers with a GELU activation between them,
% and their hidden dimension are both set to be 4 times of the input dimension.
The structures of Multi-Block Token Fusion and Spatial Token Fusion for LLaVA-1.5-7B are listed in Table~\ref{tab:arch}, respectively.

\begin{table}[ht!]
    \centering
    \vspace{-1em}
    \caption{The model structures of MBTF and STF. 
    ``act'' denotes the activation function after the convolution module.
    }
    % \renewcommand{\arraystretch}{1.5}
    % \vspace{-1em}
    \begin{subtable}[t]{0.49\linewidth}
    \resizebox{\linewidth}{!}
    {
    \begin{tabular}{c|c|c|c|c}
        \hline
        \textbf{layer} & \textbf{output size} & \textbf{kernel size} & \textbf{stride} & \textbf{act} \\
        \hline
        conv1 & $24 \times 24 \times 4096$ & $1 \times 1$ & 1 & GeLU \\
        \hline
        conv2 & $24 \times 24 \times 1024$ & $1 \times 1$ & 1 & GeLU \\
        \hline
    \end{tabular}
    }
    \caption{Multi-Block Token Fusion (MBTF)}
    \end{subtable}
    \begin{subtable}[t]{0.49\linewidth}
    \resizebox{\linewidth}{!}
    {
    \begin{tabular}{c|c|c|c|c}
        \hline
        \textbf{layer} & \textbf{output size} & \textbf{kernel size} & \textbf{stride} & \textbf{act} \\
        \hline
        conv1 & $12 \times 12 \times 4096$ & $2 \times 2$ & 2 & GeLU \\
        \hline
        conv2 & $12 \times 12 \times 16384$ & $1 \times 1$ & 1 & GeLU \\
        \hline
        conv3 & $12 \times 12 \times 4096$ & $1 \times 1$ & 1 & GeLU \\
        \hline
    \end{tabular}
    }
    \caption{Spatial Token Fusion (STF)}
    \end{subtable}
    \vspace{-1em}
    \label{tab:arch}
\end{table}

We compare our method with 6 efficient LLaVA-style methods, including LLaVA-1.5~\cite{llava}, PruMerge+~\cite{prumerge}, FastV~\cite{fastv}, LLaVolta~\cite{llavolta}, YOPO~\cite{yopo} and LLaVA-Mini~\cite{llavamini}.
All above methods use CLIP ViT-L/14 as the vision encoder, and Vicuna-1.5-7B as the LLM backbones.
For fairness comparison, all methods follow the data preparation of LLaVA-1.5~\cite{llava}.

\subsubsection{Evaluation Benchmarks}
We evaluate our method on 8 popular vision-language benchmarks, 
including GQA~\cite{gqa}, ScienceQA (SQA)~\cite{sqa}, VQAv2~\cite{vqav2}, VisWiz~\cite{viswiz}, TextVQA~\cite{textvqa}, POPE~\cite{pope}, MMBench~\cite{mmebnch} and MMBench-CN~\cite{mmebnch}.
GQA~\cite{gqa} tests fine-grained visual reasoning with multistep question-answer pairs.
SQA~\cite{sqa} with multiple choice are used to evaluate the zero-shot generalization on scientific question answering,
and we focus on the SQAI subset, which contains questions that specifically include images as part of the question context.
VQAv2~\cite{vqav2} tests visual and commonsense reasoning with open-ended questions on images.
VizWiz~\cite{viswiz} contains 8,000 images to evaluate model's zero-shot generalization on visual questions asked by visually impaired people.
TextVQA~\cite{textvqa} contains text-rich visual question answering.
POPE~\cite{pope} measures object hallucination under varying conditions, and we report the average F1 score on all conditions.
MMBench~\cite{mmebnch} and the CN version~\cite{mmebnch} evaluate a model's answer robustness with all-round shuffling on multiple choice answers.
MME-Perception~\cite{mme} evaluates model's visual perception with yes/no questions.

\begin{table*}[t!]
    \centering
    \caption{
    Performance comparison on 8 popular vision-language reasoning benchmarks.
    VQA$^{T}$ is short for TextVQA.
    ``Avg.'' denotes the average score of the previous 8 vision-language benchmarks. 
    Method marked with ``*'' is reproduced by us according to the official code.
    Our method is finetuned on full parameters of LLM, instead of LoRA.
    The best results are marked by \textbf{bold font}.
    The second-best results are marked by \underline{underline}.
    }
    \resizebox{\linewidth}{!}{
    \begin{tabular}{c|c|c|cccccccc|c}
    % \toprule
    \hline
    \textbf{Backbone} 
    & \textbf{Method} & \textbf{TFLOPs} 
    & \textbf{GQA} & \textbf{SQA} & \textbf{VQA$^{T}$} & \textbf{POPE} & \textbf{MMB} 
    & \textbf{MMB$^{CN}$} & \textbf{VQA$^{v2}$} & \textbf{VisWiz} 
    & \textbf{Avg. (\%) } \\ 
    \hline
    \multirow{9}{*}{Vicuna-1.5-7B} 
    & LLaVA-1.5 (lora)~\cite{llava2} & 7.6 
    & 63.0 & 68.4 & 58.2 & 86.4 & 66.1 
    & 58.9 & 79.1 & 47.8 & 66.0 \\
    & LLaVA-1.5 (full tuning)~\cite{llava2} & 7.6 
    & 62.0 & 66.8 & 58.2 & 85.9 & 64.3 
    & 58.3 & 78.5 & 50.0 & 65.5 \\
    \cline{2-12}
    & LLaVolta~\cite{llavolta} & 5.0 
    & 62.1 & 70.5 & 58.7 & 86.3 & 65.6 
    & 59.9 & 78.8 & 48.3 & 66.3 \\
    \cline{2-12}
    & PruMerge+~\cite{prumerge} & 1.9 & 59.3 & 68.3 & 57.1 & 84.0 & 64.9 & 53.2 & 76.8 & 49.8 & 64.2 \\
    & FastV~\cite{fastv} & 1.9 
    & 60.3 & -  & \textbf{57.7} & 83.2 & 64.3 
    & 58.0 & \underline{77.7} & \underline{50.8} & 64.6 \\
    & YOPO~\cite{yopo} & 1.9 
    & 61.6 & 69.0 & 56.3 & \textbf{86.8} & 65.5 
    & \textbf{59.6} & 78.0 & 49.9 & \underline{65.8} \\
    % & \textbf{LLaVA-Mini} & 1.9 & 60.9 & 70.4 & 57.0 & 84.4 & 65.6 & 59.8 & 77.6 & \textbf{56.2} & \textbf{67.2} & 1466.0 \\
    & LLaVA-Mini*~\cite{llavamini} & 1.9 
    & 58.4 & 67.2 & 55.2 & 83.2 & 63.2 
    & 55.3 & 76.2 & 48.2 & 63.4 \\
    & LLaVA-1.5 (AvgPool) & 1.9 
    & \textbf{62.1} & \underline{69.8} & 55.5 & \underline{86.3} & \underline{65.0} 
    & 57.9 & 77.0 & 44.0 & 64.7 \\
    % \cline{2-13}
    & \textbf{STC (ours)} & 1.9 
    & \underline{61.9} & \textbf{70.5} & \underline{57.4} & 86.0 & \textbf{66.2} 
    & \underline{58.9} & \textbf{78.1} & \textbf{51.1} & \textbf{66.3} \\
    \hline
    \end{tabular}
    }
    \vspace{-1em}
    \label{tab:main_table}
\end{table*}

\subsubsection{Implementations}
We follow LLaVA-1.5~\cite{llava} to perform data preparation and training schedule for pretraining and instruction tuning, and train the model from scratch with reduced spatial visual redundancy.
We pretrain our model on the filtered CC-595K~\cite{llava} subset for 1 epoch with learning rate of $1 \times 10^{-3}$ and batch size of 256, 
and finetune on the proposed LLaVA-Instruct-158K~\cite{llava} dataset for 1 epoch, with learning rate of $2 \times 10^{-5}$ and batch size of 128. 
The Adam optimizer is employed without weight decay, and the learning rate follows cosine schedule with warmup ratio of 3\%. 
For efficient GPU memory usage during finetuning, we utilize DeepSpeed~\cite{rasley2020deepspeed} and gradient checkpointing, without offloading. 
Additionally, bfloat16 and TensorFloat32 are enabled to strike a balance between computational speed and precision.
In the pretraining stage, we update the parameters of both MBTF and STF module, but fix the ones of LLM.
In the instruction tuning stage, we update the full parameters of LLM, MBTF and STF.
During pretraining and finetuing, the parameters of vision encoder are always frozen.
We conduct all above experiments on servers, each of which contains 8$\times$ Nvidia RTX A6000 GPUs.

\subsection{Comparison to Other Efficient LLaVA Models}

To validate the effectiveness of our method, we apply our method on LLaVA model and compare it with other efficient LLaVA models. 
As shown in Table~\ref{tab:main_table}, our method achieves the best average performance on 8 popular vision-language reasoning benchmarks among methods, which have a similar computation cost about 1.9 TFLOPs (evaluated by calflops~\cite{calflops}).
In spite of only 25\% vision tokens of the original LLaVA model used, our method even outperforms both LLaVA-1.5 (lora) and LLaVA-1.5 (full tuning), which adopt full sequence of vision tokens, by 0.3\% and 0.8\% average score, respectively.
The performance gain stems from high scores on SQA, MMB, VQAv2 and VisWiz benchmarks, especially on SQA. 
The experimental results demonstrate that our method can effectively reduce the spatial redundancy of vision tokens and accelerate the inference of LLaVA model, while maintaining comparable, even better performance.

Surprisingly, even simply applying average pooling on vision tokens (reduced to 25\% of the original one) and retraining the LLaVA model, it also acheves better performance than several other effecient LLaVA models. 
It indeed supports the excessive spatial redundancy among vision tokens of LLaVA. 

\begin{table*}[ht!]
    % \vspace{-1em}
    \centering
    \caption{The effect of fusion modules.
    All results are evaluated on the baseline of LLaVA-1.5 with Vicuna-1.5-7B.
    ``Avg.'' denotes the average score of the previous 8 vision-language benchmarks. 
    }
    \vspace{-0.5em}
    \resizebox{\linewidth}{!}{
    \begin{tabular}{c|c|cccccccc|c}
    % \toprule
    \hline
    \textbf{Method} & \textbf{TFLOPs} 
    & \textbf{GQA} & \textbf{SQA} & \textbf{VQA$^{T}$} & \textbf{POPE} & \textbf{MMB} 
    & \textbf{MMB$^{CN}$} & \textbf{VQA$^{v2}$} & \textbf{VisWiz} 
    & \textbf{Avg. (\%) }  \\ 
    \hline
    LLaVA-1.5 (baseline, full tuning)~\cite{llava2}  & 7.6 
    & 62.0 & 66.8 & 58.2 & 85.9 & 64.3 
    & 58.3 & 78.5 & 50.0 & 65.5 \\
    \textbf{MBTF}  & 7.6  
    & 63.2 & 69.3 & 58.9 & 86.8 & 65.3 
    & 59.5 & 79.4 & 50.5 & 66.6 \\
    \hline
    % \textbf{STF}  & 7.6 
    % & 60.8 & 70.8 & 55.7 & 86.3 & 64.1 
    % & 55.3 & 77.5 & 49.0 & 64.5 & 1446.2 \\
    \textbf{STF}  & 1.9
    & 61.8 & 70.8 & 56.7 & 86.3 & 64.9 
    & 57.3 & 77.8 & 49.5 & 65.6 \\
    \textbf{MBTF + STF (ours)}  & 1.9 
    & 61.9 & 70.5 & 57.4 & 86.0 & 66.2 
    & 58.9 & 78.1 & 51.1 & 66.3 \\
    \hline
    \end{tabular}
    }
    \vspace{-1em}
    \label{tab:module}
\end{table*}

\subsection{Ablation Study}
In this section, we further conduct ablation experiments to validate the effectiveness of our proposed modules.

\subsubsection{The Effect of Fusion Modules}
To substantiate the effectiveness of our fusion modules, we ablate each fusion module and compare their performance on 8 vision-language reasoning benchmarks. 

As Table~\ref{tab:module}, LLaVA-1.5 with only MBTF significantly surpasses the baseline LLaVA-1.5 by 1.1\% on the average score of 8 benchmarks. 
It implicates that features from previous layers of vision encoder contribute to the downstream vision-language tasks.
Furthermore, we only evaluate the performance of LLaVA-1.5 with only STF, which achieves comparable performance as the original LLaVA-1.5, using only 25\% vision tokens.
The results demonstrate that excessive spatial redundancy exists in the sequence of vision tokens.
Finally, we combine MBTF and STF to obtain more compact vision tokens. 
It achieves 0.7\% performance gain based on the model with only STF, but doesn't outperform the model with only MBTF.
The results are reasonable, since the model with MBTF and STF only costs 25\% FLOPs of the one with MBTF.
Overall, our proposed fusion modules can effectively improve the compactness of vision tokens and accelerate the inference without obvious performance drop.

\begin{table*}[t]
    \centering
    \caption{The effect of fusion kernel size $k$ and \#fused tokens $E$.
    All results are evaluated on the baseline of LLaVA-1.5 with Vicuna-1.5-7B.
    ``Avg.'' denotes the average score of the previous 8 vision-language benchmarks. 
    }
    \vspace{-0.5em}
    \resizebox{\linewidth}{!}{
    \begin{tabular}{c|c|c|cccccccc|c}
    % \toprule
    \hline
    \textbf{kernel size $k$} & \textbf{\#fused tokens $E$}  & \textbf{TFLOPs}
    & \textbf{GQA} & \textbf{SQA} & \textbf{VQA$^{T}$} & \textbf{POPE} & \textbf{MMB} 
    & \textbf{MMB$^{CN}$} & \textbf{VQA$^{v2}$} & \textbf{VisWiz} 
    & \textbf{Avg. (\%) } \\ 
    \hline
    1 & 1  & 7.6 
    & 62.0 & 66.8 & 58.2 & 85.9 & 64.3 
    & 58.3 & 78.5 & 50.0 & 65.5 \\
    \hline
    2 & 1 & 1.9 
    & 61.9 & 70.5 & 57.4 & 86.0 & 66.2 
    & 58.9 & 78.1 & 51.1 & 66.3 \\
    2 & 2 & 3.8 
    & 62.7 & 69.1 & 56.2 & 86.2 & 65.6 
    % & 58.8 & 77.9 & 46.7 & 65.4 & 1476.5 \\
    & 58.8 & 77.9 & 51.1 & 66.0 \\
    \hline
    4 & 4 & 1.9 
    & 61.2 & 69.4 & 52.0 & 84.8 & 64.2 
    & 57.2 & 77.1 & 42.3 & 63.5 \\
    4 & 8 & 3.8 
    & 60.4 & 70.1 & 53.2 & 84.6 & 64.9 
    & 57.5 & 77.2 & 46.8 & 64.3 \\
    \hline
    8 & 16 & 1.9 
    & 58.6 & 69.2 & 49.8 & 83.3 & 63.4
    & 54.4 & 75.8 & 42.5 & 62.1 \\
    8 & 32 & 3.8 
    & 59.2 & 68.6 & 48.9 & 83.5 & 62.7 
    & 54.1 & 75.6 & 39.2 & 61.5 \\
    \hline
    \end{tabular}
    }
    \label{tab:fusion}
    \vspace{-1em}
\end{table*}

\subsubsection{The Effect of Fusion Parameters}

To explore optimal hyperparameters for token fusion, we compare the performance of our method under different kernel size $k$ and different numbers of fused tokens $E$.
As results reported in Table~\ref{tab:fusion}, our method with $k=2$ and $E=1$ achieves the best performance. 
We find that more redundancy of vision tokens can not improve the performance of LLaVA model.
Our method with $k=2$ and $E=2$ also outperforms the original LLaVA, where our method only uses 50\% tokens of the original one.
However, with the increment of kernel size, the performance also obviously drops.
% We speculate that due to the large number of fusion module parameter, the model overfits the dataset. 
We speculate the potential reason as follows.
The number of SFC parameters increases with kernel size, and they are randomly initialized.
Due to the deficit of training data, the model overfits the dataset and doesn't achieve better performance.

\begin{table*}[ht!]
    % \vspace{-1em}
    \centering
    \caption{The effect of fusion strategies.
    All results are evaluated on the baseline of LLaVA-1.5 with Vicuna-1.5-7B.
    ``Avg.'' denotes the average score of the previous 8 vision-language benchmarks. 
    }
    \vspace{-0.5em}
    \resizebox{\linewidth}{!}{
    \begin{tabular}{c|c|cccccccc|c}
    % \toprule
    \hline
    \textbf{Method} & \textbf{TFLOPs} 
    & \textbf{GQA} & \textbf{SQA} & \textbf{VQA$^{T}$} & \textbf{POPE} & \textbf{MMB} 
    & \textbf{MMB$^{CN}$} & \textbf{VQA$^{v2}$} & \textbf{VisWiz} 
    & \textbf{Avg. (\%) }  \\ 
    \hline
    AvgPool & 1.9 
    & 62.1 & 69.8 & 55.5 & 86.3 & 65.0
    & 57.9 & 77.0 & 44.0 & 64.7 \\
    TokenConcat & 1.9
    & 62.5 & 69.7 & 56.1 & 85.9 & 66.2 
    & 59.2 & 77.8 & 46.3 & 65.4 \\
    \textbf{STF (ours)}  & 1.9 
    & 61.9 & 70.5 & 57.4 & 86.0 & 66.2 
    & 58.9 & 78.1 & 51.1 & 66.3 \\
    \hline
    \end{tabular}
    }
    \vspace{-1em}
    \label{tab:stf}
\end{table*}

\subsubsection{The Effect of Fusion Strategies}

We also research different fusion strategies to reduce the length of vision tokens, which are fed into LLM.
The first strategy is AvgPool, which directly averages adjacent $2 \times 2$ vision tokens as one token.
The second strategy is TokenConcat, which simply concatenates adjacent $2 \times 2$ vision tokens as one token.
The results of these strategies are reported in Table~\ref{tab:stf}.
TokenConcat outperforms AvgPool by 0.7\% average score on 8 vision-language benchmarks, yet inferior to our method.
Compared to AvgPool, TokenConcat can reduce the information loss of token fusion, thus achieving better performance.
Moreover, our method can adaptively fuse the vision tokens, thereby outperforming both AvgPool and TokenConcat.

\begin{figure*}[t]
	\centering
    % \vspace{-0.5em}
	\includegraphics[width=\linewidth]{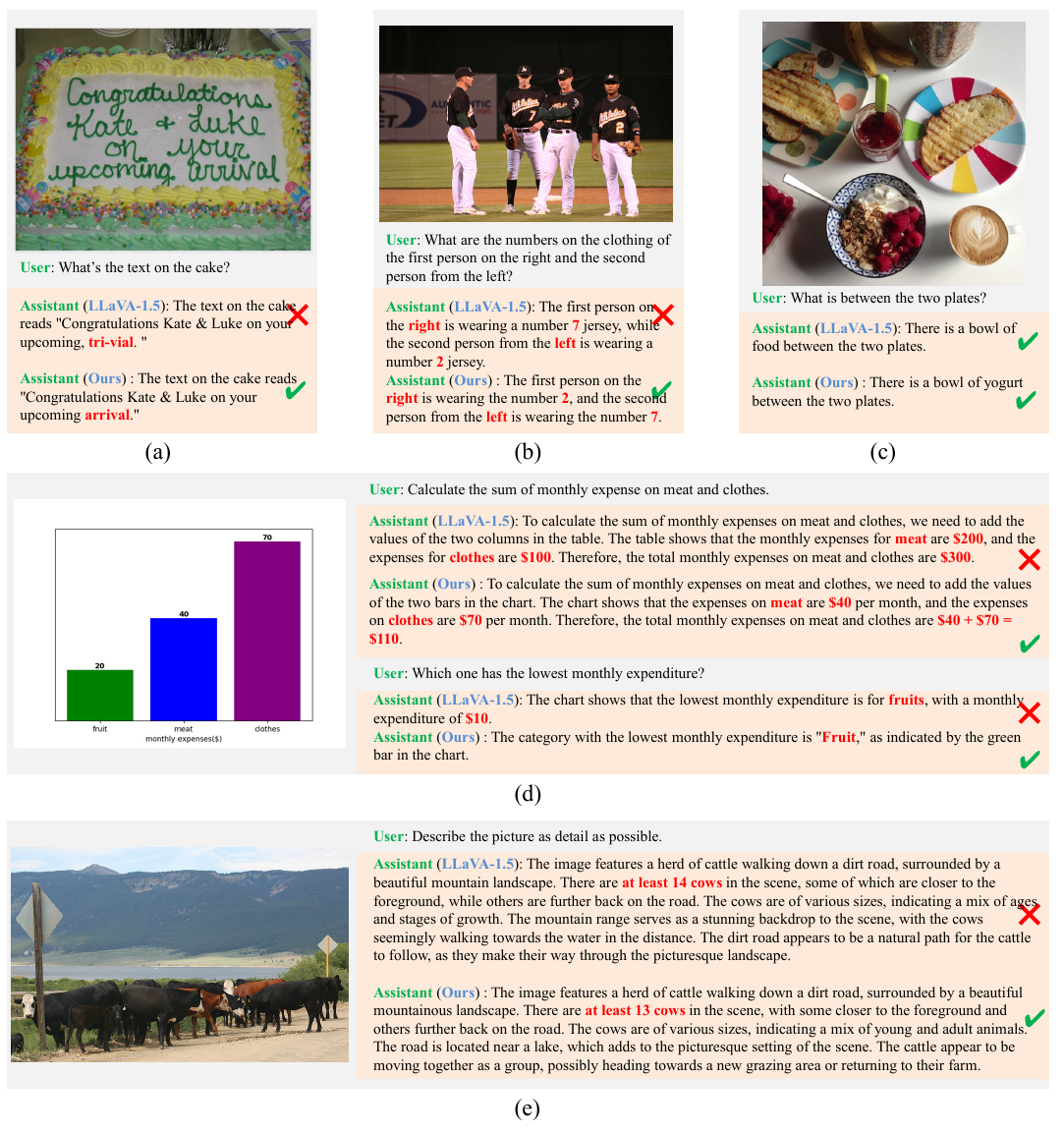}
    % \vspace{-0.5em}
    \caption{Case study of LLaVA-1.5 and our proposed method.
	}
	\label{fig:cases}
    % \vspace{-0.5em}
    \vspace{-2em}
\end{figure*}

\subsection{Cases Study}

To better understand the property of our method, we present some specific cases.
As shown in Figure~\ref{fig:cases}, we compare our method with the baseline LLaVA-1.5 with Vicuna-1.5-7B. 
In Figure~\ref{fig:cases} (a), our method successfully recognize the last word "arrival", but LLaVA-1.5 mistakes it as "tri-vial".
This case shows that our method can well recognize the blurred image information by understanding the context of images.
In Figure~\ref{fig:cases} (d), LLaVA-1.5 fails to recognize the number of bars, but understands the expenditure for fruits is lowest. 
On the contrary, our method smoothly identify both the number of bars and the lowest expenditure.
Our method also achieves superior performance on the counting of cows in Figure~\ref{fig:cases} (d).
The above cases reveal that our method can better understand image details, in spite of fewer vision tokens used. 
Overall, the above experimental results demonstrates that our method can effectively reduce the spatial redundancy, while maintaining the vision-language reasoning capabilities. 
\section{Conclusion, Limitations and Future Work}

In this paper, we propose a novel token fusion method to reduce vision tokens fed into LLM, thereby accelerating the inference of LMM.
To this end, we introduce Multi-Block Token Fusion and Spatial Token Fusion module to fuse multi-granularity representations of vision encoder and reduce spatial redundancy. 
The experimental results demonstrate that our method with only 25\% vision tokens achieves comparable or even superior performance to the baseline. 
It indeed supports that our method can effectively reduce the redundancy of vision token sequence, while maintaining the information of original images.

In spite of encouraging overall performance, our method still presents inferior results on partial vision-language benchmarks.
In future work, we plan to explore more efficient and effective strategy to reduce potential information loss during the fusion of vision tokens and further improve their compactness.

% In future work, we plan to explore more efficient token fusion and reduction strategy to further accelerate the inference of LMMs. 
% We believe that more high-quality pretraining and instruct tuning data can also effectively reduce the risk of overfitting in the current model and support larger convolution kernels for token fusion, thus improving the compactness of our vision tokens.

% \clearpage

% \section*{References}
{
\small
\bibliographystyle{plain}
% \bibliography{mybib}
\bibliography{paper.bbl}
}

\end{document}